%% file: egpaper.tex
\documentclass[10pt,twocolumn,letterpaper]{article}

\usepackage{wacv}
\usepackage{times}
\usepackage{epsfig}
\usepackage{graphicx}
\usepackage{amsmath}
\usepackage{amssymb}
\usepackage[]{algorithm2e}
\usepackage{setspace}

\wacvfinalcopy 

\ifwacvfinal
\def\assignedStartPage{9876} 
\fi

\ifwacvfinal
\usepackage[breaklinks=true,bookmarks=false]{hyperref}
\else
\usepackage[pagebackref=true,breaklinks=true,colorlinks,bookmarks=false]{hyperref}
\fi

\ifwacvfinal
\else
\fi

\begin{document}

\title{Weight and Gradient Centralization in Deep Neural Networks}

\author{Wolfgang Fuhl\\
University of T\"ubingen\\
Sand 14, 72076 T\"ubingen, Germany\\
{\tt\small wolfgang.fuhl@uni-tuebingen.de}
\and
Enkelejda Kasneci\\
University of T\"ubingen\\
Sand 14, 72076 T\"ubingen, Germany\\
{\tt\small enkelejda.kasneci@uni-tuebingen.de}
}

\maketitle

\begin{abstract}
   Batch normalization is currently the most widely used variant of internal normalization for deep neural networks. Additional work has shown that the normalization of weights and additional conditioning as well as the normalization of gradients further improve the generalization. In this work, we combine several of these methods and thereby increase the generalization of the networks. The advantage of the newer methods compared to the batch normalization is not only increased generalization, but also that these methods only have to be applied during training and, therefore, do not influence the running time during use. Link to CUDA code \url{https://atreus.informatik.uni-tuebingen.de/seafile/d/8e2ab8c3fdd444e1a135/}
\end{abstract}

\section{Introduction}
\input{introduction}

\section{Related Work}
\input{relatedwork}

\section{Method}
\input{mehod}

\input{evaluation}

\section{Limitations}
\input{limitations}

\section{Conclusion}
\input{conclusion}


{\small
\bibliographystyle{ieeefullname}
\bibliography{egbib}
}

\end{document}

%% file: introduction.tex
Deep neural networks (DNN)~\cite{lecun1998gradient} are currently the most successful machine learning method and owe their recent progress to the steadily growing data sets~\cite{krizhevsky2012imagenet}, improvements in massively parallel architectures~\cite{kirk2007nvidia}, high-speed bus systems such as PCIe, optimization methods~\cite{qian1999momentum,kingma2014adam}, new training techniques~\cite{goodfellow2014generative,kingma2015variational}, validation~\cite{ICMV2019FuhlW,NNVALID2020FUHL}, and the regularly growing fields of application like eye tracking~\cite{WF042019,VECETRA2020,NNETRA2020} for pupil~\cite{WTCDAHKSE122016,WTCDOWE052017,WDTTWE062018,VECETRA2020,CORR2017FuhlW1,ETRA2018FuhlW} or eyelid extraction~\cite{WTDTWE092016,WTDTE022017,WTE032017}, semantic segmentation~\cite{ICCVW2019FuhlW,CAIP2019FuhlW,ICCVW2018FuhlW} or gesture recognition~\cite{UMUAI2020FUHL}. These advances in technology make it possible to train deep neural networks on huge datasets like ImageNet~\cite{krizhevsky2012imagenet}, however, further techniques had to be introduced to prevent the gradients from becoming too small~\cite{he2016deep}. The normalization of the data~\cite{ioffe2015batch} has a huge impact on the generalization of large networks. Generalization alone is not the only quality feature of a good learning process of neural networks. Another important point is the acceleration of the learning process and the resource-saving~\cite{AAAIFuhlW} use of the techniques. This is due to the fact that the most successful architectures already have an intrinsically high resource requirement and additional techniques to improve generalization can, therefore, only use a small number of supplementary resources. This can be seen very clearly when comparing the optimization techniques themselves. The most popular methods are Stochastic Gradient Decent (SGD) with momentum~\cite{qian1999momentum} and Adam~\cite{kingma2014adam} which introduces a second momentum. There are many other optimization algorithms~\cite{qian1999momentum,kingma2014adam,bottou1991stochastic,duchi2011adaptive}, but SGD and Adam are the most popular. Both methods allow batch based learning and require only a constant multiple of the gradient (for the momentum) as additional memory. Comparing this with the Levenberg-Marquardt algorithm (LM)~\cite{marquardt1944method,marquardt1963algorithm}, which was the most popular method for training neural networks for quite some time, it is noticeable that the memory consumption in the case of LM grows quadratic to the weights. This is due to the fact that the LM algorithm calculates the exact derivatives for each weight over the whole network and not only local derivatives as is the case with backpropagation. Further procedures like the residual layers~\cite{he2016deep}, weight initialization strategy~\cite{glorot2010understanding,he2015delving}, activation functions~\cite{nair2010rectified}, gradient clipping~\cite{pascanu2012understanding,pascanu2013difficulty}, algorithms for adaptive learning rate optimization~\cite{qian1999momentum,kingma2014adam}, and many more have been introduced and are subject to the same conditions of generalization improvement, training stabilization, and resource conservation.

In neural networks themselves, statistics are also collected and used to balance the forward and backward flow of data and errors. The best known method used directly on the activation of neurons is Batch Normalization (BN)~\cite{ioffe2015batch}. Other procedures that work on the activation functions are instance normalization (IN)~\cite{ulyanov2016instance,huang2017arbitrary}, layer normalization (LN)~\cite{ba2016layer} and group normalization (GN)~\cite{wu2018group}. These procedures smooth the optimization landscape~\cite{santurkar2018does} and lead to an improvement of the generalization. The disadvantages of BN are that it continues to process the data in the neural network as an independent layer even after training and that it must be applied to a relatively large batch size. To avoid these disadvantages, weight normalization (WN)~\cite{salimans2016weight,huang2017centered} and weight standardization (WS)~\cite{qiao2019weight} were introduced. These must only be applied during training and are independent of the batch size. WN limits the weight vectors via different standards whereas WS normalizes the weight vectors via the mean and standard deviation. A newer technique that works only on the gradient is Gradient Centralization (GC)~\cite{yong2020gradient}, which subtracts the mean value from the gradient. All these advanced techniques smooth the error space and lead to a faster and, typically, better generalization of neural networks.

In this work, we deal with these extended techniques and seek to find a good combination of the methods. In our evaluation, it has been shown that the combination of the filter mean subtraction and the gradient mean subtraction in union is very effective for different networks. We have also tested other combinations and found that many also depend on batch normalization. Our main contributions are as follows:

\begin{description}
	\setlength\itemsep{-0.5em}
	\item[1] The combination of mean gradient and mean filter subtraction
	\item[2] Publicly available CUDA implementations
	\item[3] Description of the integration into the back propagation algorithm
	\item[4] A comprehensive comparison with advanced techniques
\end{description}

%% file: relatedwork.tex
In this section, we describe the related work based on three groups. The first group is the manipulation of the data after the activation functions, which has the disadvantage that the activation functions have to be executed in the later application of the model. The second group is the manipulation of the weights during training. Here, the weights can be standardized or otherwise restricted. The last group is the manipulation of the gradients. In this instance, after each back propagation, normalizations and restrictions are applied to the gradients before they are being used to change the weights.

\subsection{Manipulation of the output of the activations}
This type of normalization is the most common use of internal manipulation in DNNs today. In batch normalization (BN)~\cite{ioffe2015batch}, the mean value and standard deviation are calculated over several batches and used for normalization. This gives the output of neurons after the activation layer a mean value of zero and uniform variance. With group normalization GN~\cite{wu2018group}, groups are formed over which normalization is performed unlike BN where normalization occurs over the number of copies in a batch, this eliminates the need for large batches, which is the case with BN. Other alternatives are instance normalization IN~\cite{ulyanov2016instance,huang2017arbitrary} and layer normalization LN~\cite{ba2016layer}. For IN, each specimen is used individually for the calculation of the mean and standard deviation, and for LN, the individual layers. IN and LN have been successfully used for recurrent neural networks (RNN)~\cite{schuster1997bidirectional}. However, all these methods have the disadvantage that normalization has to be applied even after training.

\subsection{Manipulation of the weights of the model}
In weight normalization (WN)~\cite{salimans2016weight,huang2017centered}, the weights of the neural network are multiplied by a constant divided by the Euclidean distance of the weight vector of a neuron. This decouples the weights with respect to their length, thereby accelerating the training. An extension of this method is weight standardization (WS)~\cite{qiao2019weight}, which does not require a constant, but calculates the mean and standard deviation thus normalizing the weights. Like the previous manipulation methods, this method smooths the error landscape, which speeds up training and levels the generalization of the final model. An advantage of these methods is that they only have to be applied during training and not in the final model. These methods, however, have a limitation and that is the fine-tuning of neural networks. If the original model was not trained with a weight normalization, these methods cannot be used for fine-tuning without creating a high initial error on the model. This is due to the fact that the restrictions and norms for the original model's weights most likely do not apply.

\subsection{Manipulation of the gradient after back propagation}
Another very common technique is gradient manipulation over the first ~\cite{qian1999momentum} and second moment~\cite{kingma2014adam}. This gradient impulse allows neural networks to be trained in a stable way without the gradients exploding, which is interpreted as a damped oscillation. The second momentum leads, in most cases, to a faster generalization, but the model's final performance is usually slightly worse when compared to training with only the first momentum. These moments are moving averages which are formed over the calculated gradients and represent a pre-determined portion of the next weight adjustment. An advanced method in this area is gradient clipping~\cite{pascanu2012understanding,pascanu2013difficulty} wherein randomly selected gradients are set to zero or a small random value is added to each gradient. Another technique is to project the gradients onto subspaces~\cite{gupta2018cnn,larsson2017projected,cho2017riemannian}. Here, for example, the Riemannian approach is used to map the gradients onto a Riemannian manifold. After the mapping, the gradients are used to adjust the weights. Finally, in \cite{yong2020gradient} a very simple procedure was presented which subtracts the current mean value of the gradients in addition to the moments.

%% file: mehod.tex
Since our approach is a combination of several previously published approaches (Weight mean subtraction and gradient centralization), we proceed as follows in this section: we formally describe the already published methods and introduce a naming convention, which we use later in the evaluation. This should make it easier for the reader to evaluate the effectiveness of different methods. In the following, we will refer to operations on the data in forward propagation as $F_{s,c,y,x}$ where $s$ is the sample, $c$ is the channel and $y$, $x$ is the spatial position in the data. For the weights we use $W_{out,in,y,x}$. Where $out$ is the output channel, $in$ is the input channel and $y$, $x$ is the spatial position for fully connected layers. In the case of convolution layers, $y$, $x$ is the position in the two-dimensional convolution mask, which together with $in$ defines the convolution tensor. To manipulate the gradient we use $ \Delta W_{out,in,y,x}$ with the same indices as we used for the weights ($W_{out,in,y,x}$). Since a normalization can be applied not only to the data and the gradient, but also to the back propagated error, the error is denoted by $E_{s,c,y,x}$ where the indices are the same as the indices of the forward propagated data ($F_{s,c,y,x}$).

\subsection{Weight normalization}
In this section, the equations used for weight normalization are presented. In all equations, $j$ represents the axis to which the normalization was performed orthogonally. This means that we have calculated a separate mean value, standard deviation or Euclidean distance for each index of $j$.

\begin{equation}
W_{j,in,y,x} = W_{j,in,y,x} * \frac{k}{||W_{j,in,y,x}||}
\label{eq:WN}
\end{equation}

In Equation~\ref{eq:WN}, the weight normalization ~\cite{salimans2016weight,huang2017centered} is described (WN). This normalizes each weight in a tensor with the ratio of a constant $k$ (in our experiments $1$) divided by the Euclidean distance of the tensor.

\begin{equation}
W_{j,in,y,x} = W_{j,in,y,x} - \overline{W}_{j,in,y,x}
\label{eq:WC}
\end{equation}

Since the pure normalization over the mean value of the tensor of the weights has no separate designation, we use WC in our work. WC is defined in Equation~\ref{eq:WC} and calculates a separate mean value for each weight tensor and subtracts it from each weight.

\begin{equation}
W_{j,in,y,x} = \frac{W_{j,in,y,x}-\overline{W}_{j,in,y,x}}{std(W_{j,in,y,x})}
\label{eq:WS}
\end{equation}

The final normalization of the weights is the weight standardization~\cite{qiao2019weight} which is defined in Equation~\ref{eq:WS}. Here, as in WC, the mean value is subtracted and each weight of a tensor is also divided by the standard deviation.

\subsection{Gradient normalization}
In this section, the gradient normalization is introduced. Modern optimizers already use moving averaging with momentum~\cite{qian1999momentum,kingma2014adam}. We also think that the authors exploring gradient centralization~\cite{yong2020gradient} already tried different approaches like the standardization. We only present the recently published approach here. Also, as in the weight normalization section, $j$ corresponds to $j$ against the axis along which orthogonal normalization is performed.

\begin{equation}
\Delta W_{j,in,y,x} = \Delta W_{j,in,y,x} - \overline{\Delta W}_{j,in,y,x}
\label{eq:GC}
\end{equation}

As can be seen in Equation~\ref{eq:GC}, the mean value is subtracted from each gradient tensor. The mean value is recalculated for each output layer.

\subsection{Data normalization}
In this section, we briefly describe the different data normalizations. In our analysis, we only used batch normalization~\cite{ioffe2015batch}. In this section, $j$ as well as $j_1$ and $j_2$ (in case of instance normalization) stand for the axis or plane to which the normalization is orthogonal. Since scal and shift is learned in data manipulation, we denote them with $\gamma$ and $\beta$ respectively. 

\begin{equation}
F_{s,j,y,x} = \gamma*(\frac{F_{s,j,y,x}-\overline{F}_{s,j,y,x}}{std(F_{s,j,y,x})}) + \beta
\label{eq:BN}
\end{equation}

Equation~\ref{eq:BN} describes the batch normalization\cite{ioffe2015batch}. As mentioned above, $\gamma$ and $\beta$ are the scale and shift parameters which are learned during training. Since $j$ is on the second index, each channel has its own average and standard deviation.

\begin{equation}
F_{j,c,y,x} = \gamma*(\frac{F_{j,c,y,x}-\overline{F}_{j,c,y,x}}{std(F_{j,c,y,x})}) + \beta
\label{eq:LN}
\end{equation}

Equation~\ref{eq:LN} is the layer normalization~\cite{ba2016layer}. Compared to batch normalization~\cite{ioffe2015batch}, layer normalization is the normalization of the samples in a batch. This means that each sample has its own average and standard deviation.

\begin{equation}
F_{j_1,j_2,y,x} = \gamma*(\frac{F_{j_1,j_2,y,x}-\overline{F}_{j_1,j_2,y,x}}{std(F_{j_1,j_2,y,x})}) + \beta
\label{eq:IN}
\end{equation}

In the case of instance normalization~\cite{ulyanov2016instance,huang2017arbitrary}, each sample is normalized on its own. Equation~\ref{eq:IN} describes this procedure. It does not normalize along an axis like the other methods, but each sample and each channel separately. 

The only approach still missing is group normalization~\cite{wu2018group}. Here groups are formed between the individual instances, which have their own mean values and standard deviations. Since we cannot simply describe this with our annotation, the equation for the group normalization~\cite{wu2018group} is not included in this paper.

\subsection{Error normalization}
Inspired by the data normalization, we have also done some small evaluations regarding error normalization as a separate normalization approach. For this purpose, we evaluated the standardization as well as the simple mean value subtraction. The simple mean subtraction is based on the fact that, in the case of weight normalization, the simple mean has proven to be very effective. In our simple implementations we did not use the scale and shift ($\gamma$, $\beta$) parameters and applied the normalization directly.

\begin{equation}
E_{s,j,y,x} = \frac{E_{s,j,y,x}-\overline{E}_{s,j,y,x}}{std(E_{s,j,y,x})}
\label{eq:EBN}
\end{equation}

\begin{equation}
E_{j,c,y,x} = \frac{E_{j,c,y,x}-\overline{E}_{j,c,y,x}}{std(E_{j,c,y,x})}
\label{eq:ELN}
\end{equation}

The Equations~\ref{eq:EBN} and \ref{eq:ELN} describe error normalization along the channels and samples. The procedure is the same as for batch normalization~\cite{ioffe2015batch} and layer normalization~\cite{ba2016layer}. As you can see in the equations, we have omitted the learned $\gamma$ and $\beta$ parameters and the remainder of the equations are the same. Thus, the standard deviation and the mean value are calculated in each iteration and normalization is performed by subtracting the mean value and dividing by the standard deviation.

\begin{equation}
E_{s,j,y,x} = E_{s,j,y,x}-\overline{E}_{s,j,y,x}
\label{eq:EB}
\end{equation}

\begin{equation}
E_{j,c,y,x} = E_{j,c,y,x}-\overline{E}_{j,c,y,x}
\label{eq:EL}
\end{equation}

For the two other Equations~\ref{eq:EB}, \ref{eq:EL}, we calculated only the average value over the samples or the channels and subtracted it. This was recalculated accordingly in each iteration. An overview of the abbreviations used in the rest of the document is shown in Table~\ref{tbl:nmconv}.

\begin{table}
	\centering
	\setlength\tabcolsep{0.2em}
	\caption{The used naming convention for our evaluation.}
	\label{tbl:nmconv}
	\begin{tabular}{c|ccccccccccc}
		\textbf{Name} & WN & WC & WS & GC & BN & LN & IN & EBN & ELN & EB & EL \\ 
		\textbf{Eq.} & \ref{eq:WN} & \ref{eq:WC} & \ref{eq:WS}  & \ref{eq:GC}  & \ref{eq:BN}   & \ref{eq:LN}  & \ref{eq:IN}  & \ref{eq:EBN}   & \ref{eq:ELN} & \ref{eq:EB}  & \ref{eq:EL}    
	\end{tabular}
\end{table}

One way to include the normalizations is to add them to the back propagation workflow. This is shown in algorithm~\ref{alg:FWDBWDGRAD} were each normalization is placed in either the forward, backward, or gradient computation flow. Since batch normalization~\cite{ioffe2015batch} is a separate layer and learns the scaling and shifting, it was not inserted.

\begin{algorithm}
	\SetKwFunction{FMain}{Main}
	\SetKwProg{Fn}{Function}{ is}{end}
	\KwData{Data,Weights}
	\KwResult{Output}
	\Fn{Forward}{
		Weights=Normalize(Weights)\;
		Output=cuDNNFWD(Weights,Data)\;
	}
	\KwData{ErrorIn,Weights}
	\KwResult{ErrorOut}
	\Fn{Backward}{
		ErrorOut=cuDNNBWD(Weights,ErrorIn)\;
		ErrorOut=Normalize(ErrorOut)\;
	}
	\SetKwFunction{FMain}{Main}
	\SetKwProg{Fn}{Function}{is}{end}
	\KwData{ErrorIn,Data}
	\KwResult{Grad}
	\Fn{CompGrad}{
		Grad=cuDNNCompGrad(Data,ErrorIn)\;
		Grad=Normalize(Grad)\;
	}
	\caption{Algorithmic description of the function placement for the normalizations. Since batch normalization is implemented as its own layer with learned scaling and shifting, it is not considered in this illustration. It could be placed functionally in the forward propagation and would normalize the output.}
	\label{alg:FWDBWDGRAD}
\end{algorithm}

As you can see in Algorithm~\ref{alg:FWDBWDGRAD}, filter normalization is applied before use in the forward path and gradient normalization is applied immediately after the gradient calculation. This is because the filters must be adjusted first, otherwise the gradient will not match the weights and the weights will have no influence on the forward pass. For gradient normalization, it is applied after the calculation so that the gradients are correctly available for the weight update in the optimizer. In the case of the back propagated error, the error is normalized after the calculation of the back propagation, in which case it would, of course, also be possible to normalize the input error. However, since this is normalized in the previous layer, it is already normalized.

In this paper, we present the combination of GC~\cite{yong2020gradient} and WC, whereas WC without the standard deviation (division with the standard deviation) has, to our knowledge, never been published independently. GC~\cite{yong2020gradient} and WC can also be integrated into the optimizer itself. In the following, we give two examples: One for SGD~\cite{bottou1991stochastic} with momentum~\cite{qian1999momentum} and the other for ADAM~\cite{kingma2014adam}.

\begin{algorithm}
	\SetKwFunction{FMain}{Main}
	\SetKwProg{Fn}{Function}{ is}{end}
	\KwData{$W^{t}_{j,in,y,x}$,$\Delta W^{t}_{j,in,y,x}$,$\alpha$,$\psi$,$M^{t}_{j,in,y,x}$}
	\KwResult{$W^{t+1}_{j,in,y,x}$}
	\Fn{SGD}{
		$\Delta W^{t}_{j,in,y,x} = \Delta W^{t}_{j,in,y,x} - \overline{\Delta W}^{t}_{j,in,y,x}$\;
		$M^{t}_{j,in,y,x}=\psi*M^{t}_{j,in,y,x}+(1-\psi)*\Delta W^{t}_{j,in,y,x}$\;
		$W^{t+1}_{j,in,y,x}=W^{t}_{j,in,y,x}-\alpha*M^{t}_{j,in,y,x}$\;
		$W^{t+1}_{j,in,y,x}=W^{t+1}_{j,in,y,x}-\overline{W}^{t+1}_{j,in,y,x}$\;
	}
	\caption{Integration of the GC and WC normalization into the stochastic gradient decent optimization with momentum. The variables are weights $W^{t}_{j,in,y,x}$, gradients $\Delta W^{t}_{j,in,y,x}$, learning rate $\alpha$, momentum factor $\psi$, and momentum $M^{t}_{j,in,y,x}$. $j$ again is the index of the normalization.}
	\label{alg:SGD}
\end{algorithm}

In Algorithm~\ref{alg:SGD}, the integration of the GC and the WC normalization in combination with SGD is shown. Here the first line ($\Delta W^{t}_{j,in,y,x} = \Delta W^{t}_{j,in,y,x} - \overline{\Delta W}^{t}_{j,in,y,x}$) is the gradient centralization. Afterwards, the momentum is combined with the gradients by the factor $\psi$. In the next step, the weights are adjusted using the learning rate $\alpha$ and the mean value is subtracted from the final weights. This last step is the weight centralization ($W^{t+1}_{j,in,y,x}=W^{t+1}_{j,in,y,x}-\overline{W}^{t+1}_{j,in,y,x}$).

\begin{algorithm}
	\SetKwFunction{FMain}{Main}
	\SetKwProg{Fn}{Function}{ is}{end}
	\KwData{$W^{t}_{j,in,y,x}$,$\Delta W^{t}_{j,in,y,x}$,$\alpha$,$\psi_1$,$\psi_2$,$M^{t}_{j,in,y,x}$,$V^{t}_{j,in,y,x}$}
	\KwResult{$W^{t+1}_{j,in,y,x}$}
	\Fn{ADAM}{
		$\Delta W^{t}_{j,in,y,x} = \Delta W^{t}_{j,in,y,x} - \overline{\Delta W}^{t}_{j,in,y,x}$\;
		$M^{t}_{j,in,y,x}=\psi_1*M^{t}_{j,in,y,x}$\;
		$M^{t}_{j,in,y,x}+=(1-\psi_1)*\Delta W^{t}_{j,in,y,x}$\;
		
		$V^{t}_{j,in,y,x}=\psi_2*V^{t}_{j,in,y,x}$\;
		$V^{t}_{j,in,y,x}+=(1-\psi_2)*\Delta W^{t}_{j,in,y,x} \odot \Delta W^{t}_{j,in,y,x}$\;
		
		$\hat{M}^{t}_{j,in,y,x}=\frac{M^{t}_{j,in,y,x}}{1-\psi^t_1}$\;
		$\hat{V}^{t}_{j,in,y,x}=\frac{\psi_2*V^{t}_{j,in,y,x}}{1-\psi^t_2}$\;
		
		$W^{t+1}_{j,in,y,x}=W^{t}_{j,in,y,x}-\alpha*\frac{\hat{M}^{t}_{j,in,y,x}}{\sqrt{\hat{V}^{t}_{j,in,y,x}}+\epsilon}$\;
		$W^{t+1}_{j,in,y,x}=W^{t+1}_{j,in,y,x}-\overline{W}^{t+1}_{j,in,y,x}$\;
	}
	\caption{Integration of the GC and WC normalization into the ADAM optimization. The variables are Weights $W^{t}_{j,in,y,x}$, gradients $\Delta W^{t}_{j,in,y,x}$, learning rate $\alpha$, first order momentum factor $\psi_1$, second order momentum factor $\psi_2$, first order momentum $M^{t}_{j,in,y,x}$, and second order momentum $V^{t}_{j,in,y,x}$. $j$ again is the index of the normalization.}
	\label{alg:ADAM}
\end{algorithm}

In Algorithm~\ref{alg:ADAM} shows the integration of GC and WC in the ADAM optimization. For this, as with SGD, GC is applied first ($\Delta W^{t}_{j,in,y,x} = \Delta W^{t}_{j,in,y,x} - \overline{\Delta W}^{t}_{j,in,y,x}$). Then the new first order momentum is calculated in the following two lines using the factor $\psi_1$ ($M^{t}_{j,in,y,x}=\psi_1*M^{t}_{j,in,y,x} +(1-\psi_1)*\Delta W^{t}_{j,in,y,x}$). Subsequently, the second order momentum is calculated with the factor $\psi_2$ ($V^{t}_{j,in,y,x}=\psi_2*V^{t}_{j,in,y,x} +(1-\psi_2)*\Delta W^{t}_{j,in,y,x}) \odot \Delta W^{t}_{j,in,y,x}$). In the penultimate step, the weights are adjusted with the learning rate $\alpha$ as well as the two momentums ($W^{t+1}_{j,in,y,x}=W^{t}_{j,in,y,x}-\alpha*\frac{\hat{M}^{t}_{j,in,y,x}}{\sqrt{\hat{V}^{t}_{j,in,y,x}}+\epsilon}$). The last step is then, again, the weight centralization WC ($W^{t+1}_{j,in,y,x}=W^{t+1}_{j,in,y,x}-\overline{W}^{t+1}_{j,in,y,x}$).

%% file: evaluation.tex
\section{Neural Network Models}
\begin{figure*}[h]
	\centering
	\includegraphics[width=0.99\textwidth]{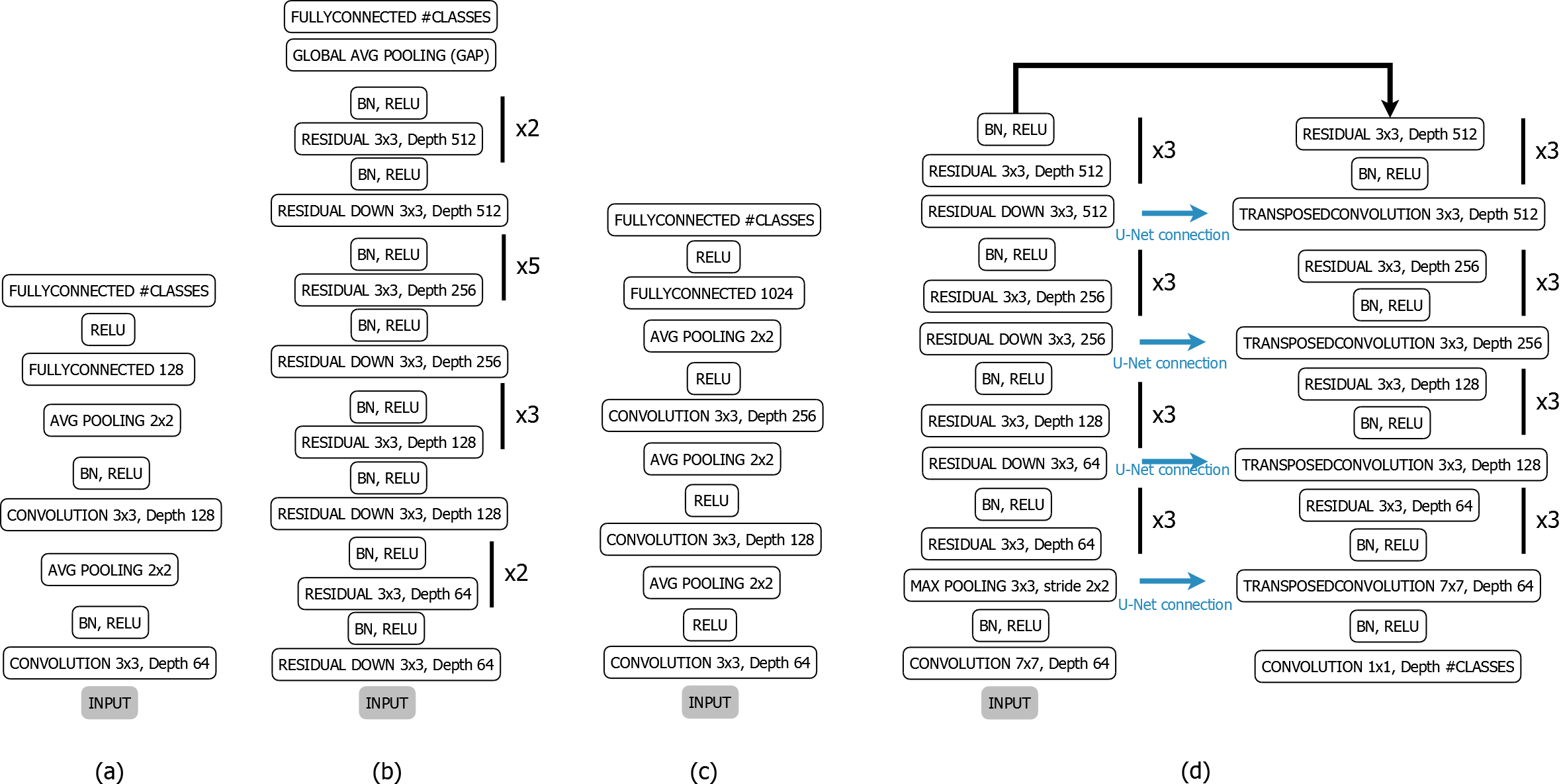}
	\caption{All used architectures in our experimental evaluation. (a) is a small neural network model with batch normalization. (b) is a ResNet-34 architecture. (c) is a small model without batch normalization. (d) is a residual network using the interconnections from U-Net~\cite{ronneberger2015u} for semantic image segmentation.}
	\label{fig:models}
\end{figure*}

Figure~\ref{fig:models} shows the architectures used in our experimental evaluation. The first model (Figure~\ref{fig:models} a)) is a small model with batch normalization. We used this model to show the impact of the different normalization approaches on small models with and without batch normalization. The second model (Figure~\ref{fig:models} b)) is a ResNet-34 and a commonly used larger deep neural network. We used it with and without batch normalization during our experiments to show the impact of the normalization approaches on residual networks. The third model (Figure~\ref{fig:models} c)) is a classical architecture for neural networks without batch normalization. This model was used to show the impact of the normalization to classical neural network architectures. The last model (Figure~\ref{fig:models} d) is a fully convolutional neural network~\cite{long2015fully}. It uses the U-connections~\cite{ronneberger2015u} to improve the result for semantic segmentation. We used this network, together with the VOC2012~\cite{pascalvoc2012} data set, in the semantic segmentation task to show the impact of the normalizations. For training and evaluation, we used the DLIB~\cite{king2009dlib} library for deep neural networks. In this library we have also integrated our normalization and the state of the art approaches against which we compare our work.

\section{Data sets}
\label{sec:datasets}
In this section, we present all used data sets, describe the used training parameters as well as the optimization techniques and the data augmentation. For a simplified reproductibility, we have limited ourselves to a minimum of data manipulation and only used public data sets. The batch size and input resolution, as well as the random weight initialization, are given too.

\textbf{CIFAR10}~\cite{krizhevsky2009learning} has 60,000 colour images each with a resolution of $32 \times 32$. The public data set has ten different classes. For training, 50,000 images are provided with 5,000 examples per class. The validation set consists of 10,000 images with 1,000 examples for each class. The task in this data set is to classify a given image to one of the ten categories.

\textit{\textbf{Training:} We used a batch size of 50 and an initial learning rate of $10^{-3}$. As optimizer, we used ADAM~\cite{kingma2014adam} with weight decay of $5*10^{-5}$, momentum one with $0.9$ and momentum two with $0.999$. As random weight initialization we used formula 16 from \cite{glorot2010understanding} and all bias terms are set to 0. For data augmentation, we cropped a $32 \times 32$ region from a $40 \times 40$ image with zero padding of the original image at the borders. In addition, we used a constant mean subtraction (mean-red $122.782$, mean-green $117.001$, mean-blue $104.298$) and division by $256.0$ for the input image. The training itself was conducted for 300 epochs whereby the learning rate was decreased by $10^{-1}$ after each 50 epochs.}

\textbf{CIFAR100}~\cite{krizhevsky2009learning} is a more difficult but similar public data set like CIFAR10 and consists of color images each with a resolution of $32 \times 32$. The task here, as in CIFAR10, is to classify the given image to one of the one hundred classes provided. The training set consists of 500 examples per class and the validation set has 100 examples per class. This means that CIFAR100 has the same number of images as CIFAR10 for training and validation, but one hundred instead of ten classes.

\textit{\textbf{Training:} We used a batch size of 50 and an initial learning rate of $10^{-1}$. As optimizer we used SGD with momentum~\cite{qian1999momentum} $0.9$ and a weight decay of ($5*10^{-4}$). For data augmentation, we cropped a $32 \times 32$ region from a $40 \times 40$ image with a zero padding of the original image at the borders. In addition, we used a constant mean subtraction (mean-red $122.782$, mean-green $117.001$, mean-blue $104.298$) and division by $256.0$ for the input image. The training itself was conducted for 300 epochs whereby the learning rate was decreased by $10^{-1}$ after each 50 epochs. For weight initialization we used formula 16 from \cite{glorot2010understanding} and all bias terms are set to 0.}

\textbf{VOC2012}~\cite{pascalvoc2012} is a publicly available data set which can be used for detection, classification and semantic segmentation. In our experiments we only used the semantic segmentation annotations as well as the semantic segmentation task. For the semantic segmentation task, a class is assigned for each output pixel. This data set has twenty different classes and each image can contain different object classes and different amounts of the same object. This also means that not every object is present in every image. The training set consists of 1,464 images with 3,507 segmented objects and the validation set has 1,449 images with a total of 3,422 segmented objects on it. The number of objects in this data set is not balanced, which increases the challenge. There is also a third data set which does not contain any annotations and can be used initially in an unsupervised fashion to have a good weight initialization. We did not use the third data set in our training nor in our evaluation.

\textit{\textbf{Training:}  The initial learning rate was set to $10^{-1}$ with a constant batch size of ten. As optimizer we used SGD with momentum~\cite{qian1999momentum} set to $0.9$ and additionally weight decay of $1*10^{-4}$. For weight initialization, we used formula 16 from \cite{glorot2010understanding} and all bias terms were set to 0. The data was augmented by cropping $227 \times 227$ regions out of the input image. In addition, we used random color offset and left right flipping of the image. Before the image was processed we subtracted a constant  mean (mean-red $122.782$, mean-green $117.001$, mean-blue $104.298$) and divided each value by $256.0$. We trained each model for 800 epochs and reduced the learning rate by $10^{-1}$ after each 200 epochs.}

\section{Evaluation}
In this section, we show the results on CIFAR10, CIFAR100 and VOC2012. We use the models from Figure~\ref{fig:models} and apply the training parameters and procedures from Section~\ref{sec:datasets}. In the first experiment, we show over which areas in the data the mean value subtraction can be used most effectively. In the following three experiments we compare the combination of GC and WC with the state of the art.

\begin{table}
	\centering
	\caption{The results for the mean subtraction normalization on CIFAR10 on different target areas for mean computation. The used model was c) from Figure~\ref{fig:models}.}
	\label{tbl:cif10wheretoapply}
	\begin{tabular}{lcc}
		\textbf{Reference area} & Target & Accuracy \\ \hline
		Baseline  & non & 84.14\% \\ \hline
		Global  & Weight  & 84.91\% \\
		Tensor  & Weight (WC) & \textbf{85.95\%} \\
		Channel & Weight  &  85.63\% \\
		Instance & Weight  & 84.37\% \\ \hline
		Global  & Gradient  & 84.01\% \\
		Tensor  & Gradient (GC~\cite{yong2020gradient}) & 84.89\% \\
		Channel & Gradient  & 84.38\% \\
		Instance & Gradient  & 83.36\% \\ \hline
		Global  & ERROR  & 79.03\% \\
		Sample  & ERROR (EL) & 72.15\% \\
		Channel & ERROR (EB) & 75.40\% \\
		Instance & ERROR  & 69.73\% 
	\end{tabular}
\end{table}

Table~\ref{tbl:cif10wheretoapply} shows the evaluation of mean subtraction on different areas of weights, gradients and back propagated errors. As can be seen, the mean subtraction on the error propagated back is not very effective because it significantly worsens the generalization of the deep neural network. This shows that error normalization without data normalization, as it happens in batch normalization~\cite{ioffe2015batch}, only brings disadvantages. For the weights and gradients, a convolutional tensor seems to be most effective for normalization. This means that each tensor is used for the mean calculation and this mean is subtracted only from this tensor. It is also clearly seen that weight normalization provides better results independent of gradient normalization with the exception of instance based normalization. In instance based normalization, an average value is calculated for every two dimensional mask and this average value is subtracted from the mask. Based on these results, we decided to define WC on the tensor and to discard the normalization of the back propagated error for further evaluations.

\begin{table}
	\centering
	\caption{Classification accuracy on the CIFAR10 data set. The first column specifies the methods, the second column the used model from Figure~\ref{fig:models}, and the third column is the classification accuracy. Models a) and b) where evaluated with and without batch normalization. For the models a) and c) we also used the normalization in the penultimate fully connected layer which is specified with the keyword \textit{fully}.}
	\label{tbl:cif10}
	\begin{tabular}{lcc}
		\textbf{Method} & Model & Accuracy  \\ \hline
		Baseline  & a & 81.87\% \\
		WN~\cite{salimans2016weight,huang2017centered} $k=1$  & a & 71.74\% \\
		WC  & a & 85.01\% \\
		WS~\cite{qiao2019weight} & a & 73.00\% \\
		GC~\cite{yong2020gradient} & a & 81.42\% \\
		WS~\cite{qiao2019weight}, GC~\cite{yong2020gradient} & a &  74.51\% \\
		WC, GC~\cite{yong2020gradient} & a & 85.75\% \\ 
		WC, GC~\cite{yong2020gradient}, fully & a & \textbf{87.07\%} \\ \hline
		Baseline, BN~\cite{ioffe2015batch}  & a & 84.67\% \\
		WN~\cite{salimans2016weight,huang2017centered} $k=1$, BN  & a & 82.95\% \\
		WC, BN~\cite{ioffe2015batch}  & a & 85.38\% \\
		WS~\cite{qiao2019weight}, BN~\cite{ioffe2015batch} & a & 79.65\% \\
		GC~\cite{yong2020gradient}, BN~\cite{ioffe2015batch} & a & 84.01\% \\
		WS~\cite{qiao2019weight}, GC~\cite{yong2020gradient}, BN~\cite{ioffe2015batch} & a &  81.01\% \\
		WC, GC~\cite{yong2020gradient}, BN~\cite{ioffe2015batch} & a & 85.48\% \\ 
		WC, GC~\cite{yong2020gradient}, BN~\cite{ioffe2015batch}, fully & a & \textbf{85.95\%} \\ \hline
		Baseline  & b & 88.35\% \\
		WN~\cite{salimans2016weight,huang2017centered} $k=1$  & b & 58.77\% \\
		WC  & b & 80.15\% \\
		WS~\cite{qiao2019weight} & b & nan \\
		GC~\cite{yong2020gradient} & b & 69.85\% \\
		WC, GC~\cite{yong2020gradient} & b & \textbf{89.61\%} \\ \hline
		Baseline, BN~\cite{ioffe2015batch}  & b & 91.00\% \\
		WN~\cite{salimans2016weight,huang2017centered} $k=1$, BN~\cite{ioffe2015batch}  & b & 61.02\% \\
		WC, BN~\cite{ioffe2015batch}  & b & 92.50\% \\
		WS~\cite{qiao2019weight}, BN~\cite{ioffe2015batch} & b & 79.83\% \\
		GC~\cite{yong2020gradient}, BN~\cite{ioffe2015batch} & b & 92.01\% \\
		WS~\cite{qiao2019weight}, GC~\cite{yong2020gradient}, BN~\cite{ioffe2015batch} & b &  79.71\% \\
		WC, GC~\cite{yong2020gradient}, BN~\cite{ioffe2015batch} & b & \textbf{92.68\%} \\ \hline
		Baseline  & c & 84.14\% \\
		WN~\cite{salimans2016weight,huang2017centered} $k=1$  & c & 83.73\% \\
		WC  & c & 85.95\% \\
		WC, fully  & c & 86.64\% \\
		WS~\cite{qiao2019weight} & c & 10.05\% \\
		GC~\cite{yong2020gradient} & c &  84.89\% \\
		GC~\cite{yong2020gradient}, fully & c &  85.37\% \\
		WS~\cite{qiao2019weight}, GC~\cite{yong2020gradient} & c &  10.72\% \\
		WC, GC~\cite{yong2020gradient} & c &  87.46\% \\
		WC, GC~\cite{yong2020gradient}, fully & c &  \textbf{87.62\%}
	\end{tabular}
\end{table}

In Table~\ref{tbl:cif10}, the results on CIFAR10 show different normalizations and the baseline, which is CNN without normalization. As you can see, the combination WC and GC can be effectively applied to all convolutions and also to the penultimate fully connected layer (indicated by the keyword \textit{fully}). This can be seen in model a) and c) from Figure~\ref{fig:models}. In model b), there is only one fully connected layer in which normalization is not effective because it generates the output. For model a) and b), we have also performed the evaluations with and without batch normalization. As you can see, the combination WC and GC works even better without batch normalization for model a). This is the best result for the model, especially together with normalization in the penultimate fully connected layer. In case of model b), the additional use of batch normalization is much better, because of the residual blocks. Therefore, for the additional evaluations, all residual blocks were evaluated with batch normalization only. Since model c) does not have an integrated batch normalization, we only evaluated without batch normalization.

The combination of WS and GC has not proven to be advantageous for all models, which is why we will not use it in further evaluations. In general, the best normalization across all evaluations on CIFAR10 is the combination of WC and GC. For residual blocks, batch normalization is added. Considering normalizations individually without batch normalization, WC is clearly the best, with GC a close second.

\begin{table}
	\centering
	\caption{Classification accuracy on the CIFAR100 data set. The first column specifies the methods, the second column the used model from Figure~\ref{fig:models} and the third column is the classification accuracy. Model a) was evaluated with and without batch normalization. For the models a) and c) we also used normalization in the penultimate fully connected layer which is specified with the keyword \textit{fully}.}
	\label{tbl:cif100}
	\begin{tabular}{lcc}
		\textbf{Method} & Model & Accuracy  \\ \hline
		Baseline  & a & 46.31\% \\
		WN~\cite{salimans2016weight,huang2017centered} $k=1$  & a & 45.93\% \\
		WC  & a & 50.19\% \\
		WS~\cite{qiao2019weight} & a & 41.19\% \\
		GC~\cite{yong2020gradient} & a & 45.55\% \\
		WC, GC~\cite{yong2020gradient} & a & 50.99\% \\ 
		WC, GC~\cite{yong2020gradient}, fully & a & \textbf{52.03\%} \\ \hline
		Baseline, BN~\cite{ioffe2015batch}  & a & 54.04\% \\
		WN~\cite{salimans2016weight,huang2017centered} $k=1$, BN  & a & 44.26\% \\
		WC, BN~\cite{ioffe2015batch}  & a & 55.01\% \\
		WS~\cite{qiao2019weight}, BN~\cite{ioffe2015batch} & a & 48.99\%  \\
		GC~\cite{yong2020gradient}, BN~\cite{ioffe2015batch} & a & 53.59\% \\
		WC, GC~\cite{yong2020gradient}, BN~\cite{ioffe2015batch} & a & 56.45\% \\ 
		WC, GC~\cite{yong2020gradient}, BN~\cite{ioffe2015batch}, fully & a & \textbf{56.78\%} \\ \hline
		Baseline, BN~\cite{ioffe2015batch}  & b & 68.99\% \\
		WN~\cite{salimans2016weight,huang2017centered} $k=1$, BN~\cite{ioffe2015batch}  & b & 52.97\% \\
		WC, BN~\cite{ioffe2015batch}  & b & 69.89\% \\
		WS~\cite{qiao2019weight}, BN~\cite{ioffe2015batch} & b & 63.52\% \\
		GC~\cite{yong2020gradient}, BN~\cite{ioffe2015batch} & b & 69.34\% \\
		WC, GC~\cite{yong2020gradient}, BN~\cite{ioffe2015batch} & b & \textbf{70.24\%} \\ \hline
		Baseline  & c & 46.31\% \\
		WN~\cite{salimans2016weight,huang2017centered} $k=1$  & c & 10.31\% \\
		WC  & c & 52.05\% \\
		WS~\cite{qiao2019weight} & c & 34.65\% \\
		GC~\cite{yong2020gradient} & c & 47.95\% \\
		WC, GC~\cite{yong2020gradient} & c & 53.16\% \\
		WC, GC~\cite{yong2020gradient},fully & c & \textbf{53.90\%} 
	\end{tabular}
\end{table}

Table ~\ref{tbl:cif100} shows the results of models a), b), and c) of Figure~\ref{fig:models} on the CIFAR100 data set. As you can see, again the combination WC and GC is the most effective. As with CIFAR10 (Table~\ref{tbl:cif10}), this applies in particular to the additional use of normalization in the last fully connected layer (Indicated by the keyword \textit{fully}). Like CIFAR10 (Table~\ref{C10}), the normalization WC always delivers better results in comparison to GC, if both normalizations are evaluated alone. However, there is a difference in the batch normalization for model a). The additional batch normalization is much more effective than model a) is without batch normalization. In all evaluations in Tables~\ref{tbl:cif10} and \ref{tbl:cif100} one also sees that the normalizations WS and WN have worsened the generalization of the model. In one case, WS even led to a NaN result.

\begin{table}
	\centering
	\caption{The average pixel accuracy classification results for different normalization methods on the VOC2012 validation set using model d) from Figure~\ref{fig:models}. We applied the normalization specified in column one to all layers except for the last convolution.}
	\label{tbl:segi}
	\begin{tabular}{lc}
		\textbf{Method} & Average Pixel Accuracy \\ \hline
		Baseline, BN~\cite{ioffe2015batch}  & 85.15\% \\
		WS~\cite{qiao2019weight}, BN~\cite{ioffe2015batch} & 81.23\% \\
	 	WN~\cite{salimans2016weight,huang2017centered} $k=1$& 75.76\% \\
		GC~\cite{yong2020gradient}, BN~\cite{ioffe2015batch} & 85.91\% \\
		WC, BN~\cite{ioffe2015batch} & 86.92\% \\
		WC, GC~\cite{yong2020gradient}, BN~\cite{ioffe2015batch} & \textbf{88.98\%}
	\end{tabular}
\end{table}

Table~\ref{tbl:segi} shows the evaluation of different normalization methods on the VOC2012 data set with model d) from Figure~\ref{fig:models}. Normalization was used in all layers except the final convolution before output. As you can see, both GC and WC improve the result significantly. In combination with the batch normalization, the result is improved by more than 3\%. This clearly shows that the combination of WC and GC can be used very effectively together with batch normalization for residual blocks. For the methods WS and WN, however, the generalization of the deep neural network is even worse.

%% file: limitations.tex
A disadvantage of WC and GC is that for residual blocks without batch normalization the results are also poor. This can be seen in Table~\ref{tbl:cif10} for model b) from Figure~\ref{fig:models}. Here you can see in the evaluations that for both, as a single normalization without batch normalization, the results are significantly worse. In combination, however, they work better than the model without normalization and without batch normalization. An advantage of the combination of WC and GC compared to batch normalization is that they only need to be used in training (see Algorithm~\ref{alg:SGD} and \ref{alg:ADAM}). For batch normalization, however, it is necessary to apply the mean subtraction, division by the standard deviation, scaling, and shift at runtime. However, since this can be calculated with a complexity linear to the input, it hardly affects the runtime.

%% file: conclusion.tex
In this work, we have shown that weight centralization is a very effective normalization method. Together with gradient centralization and, for residual networks, batch normalization, this combination exceeds the state of the art. We have also shown over which area mean subtraction is most effective. Our results were generated with four different nets on three public data sets and clearly show that the additional use of weight centralization is effective and improves the generalization of deep neural networks. Further research will evaluate the applicability of the weight and gradient normalization in the fields of gaze behaviour analysis~\cite{ROIGA2018,ASAOIB2015,DWTE022017,AGAS2018} which includes eye movement segmentation~\cite{FCDGR2020FUHL,fuhl2018simarxiv,ICMIW2019FuhlW1,ICMIW2019FuhlW2,EPIC2018FuhlW,FCDGR2020FUHL} and scan path classification~\cite{C2019,FFAO2019}.